# Exploiting Evidence-dependent Sensitivity Bounds

**Silja Renooij** and **Linda C. van der Gaag**
Department of Information and Computing Sciences, Utrecht University
P.O. Box 80.089, 3508 TB Utrecht, The Netherlands
{silja,linda}@cs.uu.nl

## Abstract

Studying the effects of one-way variation of any number of parameters on any number of output probabilities quickly becomes infeasible in practice, especially if various evidence profiles are to be taken into consideration. To provide for identifying the parameters that have a potentially large effect prior to actually performing the analysis, we need properties of sensitivity functions that are independent of the network under study, of the available evidence, or of both. In this paper, we study properties that depend upon just the probability of the entered evidence. We demonstrate that these properties provide for establishing an upper bound on the sensitivity value for a parameter; they further provide for establishing the region in which the vertex of the sensitivity function resides, thereby serving to identify parameters with a low sensitivity value that may still have a large impact on the probability of interest for relatively small parameter variations.

## 1 INTRODUCTION

The output probabilities of a probabilistic network can be highly sensitive to changes in the network's numerical parameters. As these parameters are generally estimated from (incomplete) data or assessed by human experts in the domain of application, they are inevitably inaccurate. The sensitivity of the output probabilities of the network to these inaccuracies can be evaluated by subjecting the network to a *sensitivity analysis*. Such an analysis amounts to varying the assessments for one or more of the network's parameters and investigating the effects on the probabilities of interest. Efficient algorithms are available for this purpose, that build upon the observation that the sensitivity of a probability of interest to parameter variation complies with a simple mathematical function [4, 6]. This *sensitivity function* basically expresses the output probability of interest in terms of the parameter under study [1, 4]; the constants in the function can be established from a limited number of network propagations.

Performing a sensitivity analysis is computationally feasible as long as we are interested in the sensitivity functions for a *single* output probability of interest with respect to *all* network parameters, or in the sensitivity functions for *any* number of output probabilities with respect to a *single* parameter [6]. Analysing the effects of any number of parameters on any number of output probabilities, quickly becomes infeasible, especially when taking different evidence profiles into consideration. In such cases it is very useful to have properties of sensitivity functions that are independent of the network under study, of the available evidence, or of both. Such properties can then be exploited to distinguish between parameters that may need actual analysis and parameters that cannot have any substantial influence on the output probabilities of interest and therefore do not warrant further consideration.

Several researchers addressed properties of sensitivity functions that are independent of both the network under study and the available evidence [2, 3, 8]. They showed, for example, that any sensitivity function that expresses an output probability with an original value of $p_0$ in terms of a parameter $x$ with an original value of $x_0$, is bounded by two hyperbolic functions through $(x_0, p_0)$ [8]. These bounding functions depend on $x_0$ and $p_0$ only, and are not dependent of any knowledge of the network under study. The functions further are evidence-invariant and, hence, independent of the probability of the available evidence. The bounding functions served to confirm an upper bound on the effect of infinitesimally small shifts in a parameter's original value, that is, on its sensitivity value [2, 3, 8]. The established upper bound inherits the characteristics of evidence-invariance and network-independence from the bounding functions from which it is derived.

In this paper, we study properties of sensitivity functions that depend on some knowledge of the available evidence. We show that this knowledge provides us with information about the constants of the sensitivity function, which en-ables us to provide tighter bounding functions without using any other knowledge about the network under consideration. We argue that computing these evidence-dependent bounds requires much less computational effort than establishing all complete sensitivity functions. We demonstrate that the tighter bounds can be used, for example, for providing a tighter upper bound on the sensitivity value of a parameter and for locating the region in which the vertex of a sensitivity function might reside; the latter is important to identify parameters with a low sensitivity value that may nonetheless have a high impact on the output probability of interest for non-infinitesimal parameter shifts.

The paper is organised as follows. In Section 2, we present some preliminaries on sensitivity functions. In Section 3, we derive evidence-dependent bounds on the constants of a sensitivity function. In Section 4, we exploit these bounds to bound sensitivity values and to locate the vertex of a sensitivity function, respectively; in addition, we illustrate how the derived information can be purposefully used. The paper ends with our conclusions and directions for further research in Section 5.

## 2 SENSITIVITY FUNCTIONS

Sensitivity analysis of a probabilistic network amounts to establishing, for each of the network's numerical parameters, the *sensitivity function* that expresses an output probability of interest in terms of that parameter. Let $\Pr(A = a \mid e)$, or $\Pr(a \mid e)$ for short, denote the output probability under study, where $a$ is a specific value of a variable $A$ of interest and $e$ denotes the available evidence. In addition, let $x = p(b \mid \pi)$ be the parameter under study, where $b$ is a value of some variable $B$ and $\pi$ is a combination of values for $B$'s parents. We now use $f_{\Pr(a|e)}(x)$ to denote the sensitivity function that expresses the probability $\Pr(a \mid e)$ in terms of the parameter $x$; we often omit the subscript for the function symbol $f$, as long as ambiguity cannot occur.

Any sensitivity function $f_{\Pr(a|e)}(x)$ is a quotient of two linear functions in the parameter $x$ under study [1, 4]. More formally, the function takes the form

$$f(x) = \frac{c_1 \cdot x + c_2}{c_3 \cdot x + c_4}$$

where the constants $c_j$, $j = 1, \ldots, 4$, are built from the assessments for the parameters that are not being varied[1]. The numerator of this quotient in essence describes the probability $\Pr(a, e)$ as a function of the parameter $x$ and the denominator describes $\Pr(e)$ as a function of $x$. Any sensitivity function is thus characterised by at most three constants. These constants can be feasibly determined from the network, for example by computing the probability of interest for up to three values for the parameter under study and solving the resulting system of linear equations [4], or by means of an algorithm that is closely related to junction-tree propagation [6].

[1] We assume that the parameters pertaining to the same conditional distribution as the parameter under study are co-varied proportionally [6, 8].

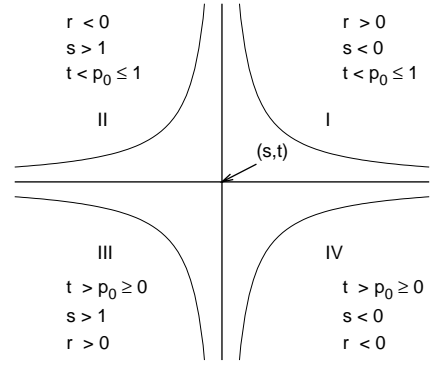

Figure 1: The possible hyperbolas and their constants (the constraints on $s$ and $t$ are specific for sensitivity functions).

A sensitivity function is either a *linear* function or a fragment of a *rectangular hyperbola*; in the remainder of this paper, we focus on hyperbolic sensitivity functions. A rectangular hyperbola takes the general form

$$f(x) = \frac{r}{x - s} + t$$

where, for a sensitivity function with $c_1, \ldots, c_4$ as before, we have that

$$s = -\frac{c_4}{c_3}, \quad t = \frac{c_1}{c_3}, \quad \text{and} \quad r = \frac{c_2 \cdot c_3 - c_1 \cdot c_4}{c_3^2}$$

The hyperbola has two branches and the two asymptotes $x = s$ and $f(x) = t$. Figure 1 illustrates the locations of the possible hyperbola branches relative to the asymptotes. For $r < 0$, the branches lie in the second (II) and fourth (IV) quadrants relative to the asymptotes; for $r > 0$, the branches are found in the first (I) and third (III) quadrants.

We observe that, in a sensitivity function, both $x$ and $f(x)$ represent probabilities and hence $x \in [0, 1]$ and $f(x) \in [0, 1]$; in the sequel we will refer to the two-dimensional space of feasible values for $x$ and $f(x)$ as the *unit window*. Since any sensitivity function is continuous and well-defined in the unit window, it is a fragment of just one of the four possible branches reviewed above. The vertical asymptote $x = s$ therefore lies either to the left of $x = 0$ (for first- and fourth-quadrant functions) or to the right of $x = 1$ (for second- and third-quadrant functions); the horizontal asymptote $f(x) = t$ either lies below 1 (for first- and second-quadrant functions) or above 0 (for third- and fourth-quadrant functions).

The sensitivity function $f(x)$ captures the change in the output probability of interest that is occasioned by a shift in the parameter $x$ under study. The effect of an infinitesi-mally small shift is captured by the value $f'(x_0)$ of the first derivative of the function at the original value $x_0$ of the parameter; the absolute value of $f'(x_0)$ is called the *sensitivity value* of the parameter for the output probability [7]. For establishing the effect of larger shifts the sensitivity value of a parameter may no longer suffice, as the impact of a larger shift is strongly dependent upon the location of the *vertex* of the sensitivity function [9]. The vertex of a hyperbola branch is the point where the absolute value of the first derivative equals 1; it is equal to one of the four points $(s \pm \sqrt{|r|}, t \pm \sqrt{|r|})$, depending on the branch's quadrant. The vertex of a hyperbolic sensitivity function may or may not lie within the unit window. If it lies outside the window, then high sensitivity values are unlikely, regardless of the original value of the parameter under study. A vertex within the unit window basically marks the transition from original parameter values with a high sensitivity value to parameter values with a low sensitivity value, or vice versa. Parameters that have a small sensitivity value, yet whose original value lies close to the vertex, may thus show considerable effects upon variation.

## 3  BOUNDING THE FUNCTIONS

A hyperbolic sensitivity function in essence is defined by the values for the three constants $r$, $s$ and $t$ reviewed above. In previous work, we studied properties of such a sensitivity function that are network independent as well as independent of the available evidence, yet build upon the actual value $p_0$ of the output probability of interest and the original value $x_0$ of the parameter under study. We established that all possible sensitivity functions through $(x_0, p_0)$ are bounded by an increasing hyperbola branch and a decreasing hyperbola branch [8]; as an illustration, Figure 2 depicts the bounding branches for $(x_0, p_0) = (0.1, 0.6)$. From $r = (x - s) \cdot (f(x) - t)$, we observe that the range of all possible sensitivity functions through $(x_0, p_0)$ is defined by just the values of $s$ and $t$: for any given $x_0$ and $p_0$, the values of $s$ and $t$ uniquely determine the value of $r$. Fixing one of $s$ and $t$ to a specific value now serves to reduce the range of possible functions through $(x_0, p_0)$, and allows for $s$- or $t$-specific properties to emerge. Knowledge of these properties can then be exploited to distinguish between the parameters for which an actual analysis may be of interest and the parameters which can be further disregarded.

In this paper, we study properties that emerge from fixing the constant $s$ to a specific value. Note that fixing $s$ means that the sensitivity functions under consideration all have their vertical asymptote at the same position. From this observation, we can establish tighter bounds on the possible functions through $(x_0, p_0)$. The reason for choosing to fix $s$ rather than $t$ is that $s$ in essence is related to the probability of the available evidence only and is independent of the output probability of interest; we will argue in Section 3.2 that, if necessary, the value of $s$ can be computed quite efficiently from a network under study. Since the constant $s$ is easily computed for *all* parameters at once, any $s$-specific properties can be immediately projected onto the network under consideration. In Section 4 we will give some examples of how these properties can be purposefully used.

### 3.1  ESTABLISHING TIGHTER BOUNDS

To provide for studying all possible sensitivity functions that pass through the point $(x_0, p_0)$ and have their vertical asymptote at the same position, we begin by defining the subspace $\mathcal{S}$ of all points in three-dimensional $(s, t, r)$-space that capture a hyperbolic sensitivity function. We then define, again within $(s, t, r)$-space, the surface that captures any rectangular hyperbola through the point $(x_0, p_0)$. The intersection of the subspace $\mathcal{S}$ and this surface then gives us all combinations of values for $s$, $t$ and $r$ that describe a hyperbolic sensitivity function through $(x_0, p_0)$.

Any hyperbolic sensitivity function $f(x)$ is continuous and well-defined in the unit window, and therefore adheres to $0 \leq f(x) \leq 1$ for all $0 \leq x \leq 1$. The subspace within $(s, t, r)$-space that is defined by this inequality, is delimited by *four* surfaces. For example, $f(0) = 1$ corresponds with the surface
$$r = (x - s) \cdot (f(x) - t) = -s \cdot (1 - t)$$
Note that the signs of $s$ and $t$ determine whether the surface is an upper or a lower bound on the values allowed for $r$. The subspace $\mathcal{S}$ of all combinations of values for $s$, $t$ and $r$ that define a hyperbolic sensitivity function, now is delimited by the following (intersecting) surfaces:

$$\begin{aligned} A: &\quad r = s \cdot (t-1) \\ B: &\quad r = t \cdot (s-1) \\ C: &\quad r = s \cdot t \\ D: &\quad r = (t-1) \cdot (s-1) \end{aligned}$$

where the surfaces $A$ and $B$ intersect for $t = s$ and the surfaces $C$ and $D$ intersect for $t = 1 - s$; the surfaces $B$ and $C$ intersect for $t = 0$ and the surfaces $A$ and $D$ intersect for $t = 1$. Figure 3 illustrates the subspace $\mathcal{S}$, where the valid combinations of values for $s$, $t$ and $r$ lie within the two regions bounded by the four surfaces. We observe that for $s > 1$ and $t \leq 0$, an upper bound on the value of $r$ is given by the surface $C$; a lower bound is given by the surface $D$. For $s > 1$ and $t \geq 0$, an upper bound on $r$ is given by the surface $B$; a lower bound is given by $A$. Similarly, for $s < 0$, if $t \leq 1$, an upper bound on the value of $r$ is given by the surface $A$ and if $t \geq 1$ an upper bound is found from $D$; the lower bound is given by the surface $B$ for $t \leq 0$ and by the surface $C$ for $t \geq 0$.

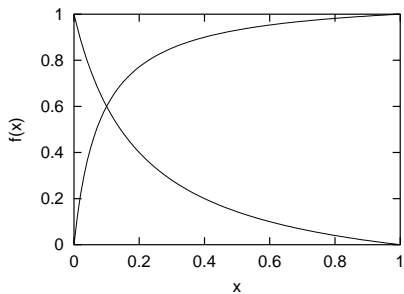

Figure 2: The general bounds on all possible sensitivity functions through $(0.1, 0.6)$.

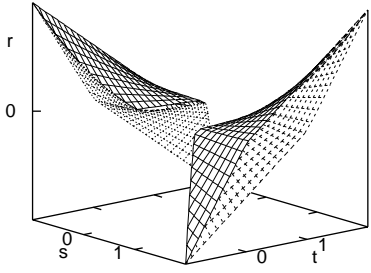

Figure 3: The subspace $\mathcal{S}$ with all $(s, t, r)$-combinations that define a hyperbolic sensitivity function.

Having defined the subspace $\mathcal{S}$ of all combinations of values for $s$, $t$ and $r$ that capture a hyperbolic sensitivity function, we now define the surface $E$ of combinations that capture a rectangular hyperbola through the point $(x_0, p_0)$ within the unit window:

$$E: \quad r = (x_0 - s) \cdot (p_0 - t)$$

The surface is depicted in Figure 4. Note that for any point on the surface $E$, the values of $s$ and $t$ uniquely determine the value of $r$.

The space of all combinations of values for the constants $s$, $t$ and $r$ that define a sensitivity function through the point $(x_0, p_0)$, now is characterised by the points from the subspace $\mathcal{S}$ that lie on the surface $E$. To establish these combinations, we determine the intersections of the surface $E$ with the four surfaces that delimit the subspace $\mathcal{S}$. The surfaces $E$ and $A$, for example, intersect at any point where $(x_0 - s) \cdot (p_0 - t) = s \cdot (t - 1)$, which results in the line $t = p_0 + (1 - p_0) \cdot \frac{s}{x_0}$. Writing $t_V$ for the values of $t$ for which the surfaces $V$ and $E$ intersect, we thus find the

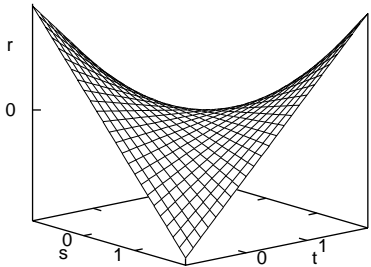

Figure 4: The surface $E$ with all $(s, t, r)$-combinations that result in $f(x_0) = p_0$.

following intersecting lines:

$$t_A = p_0 + (1 - p_0) \cdot \frac{s}{x_0}$$
$$t_B = \frac{p_0 \cdot (x_0 - s)}{x_0 - 1}$$
$$t_C = p_0 \cdot (1 - \frac{s}{x_0})$$
$$t_D = \frac{s \cdot (1 - p_0) + p_0 \cdot x_0 - 1}{x_0 - 1}$$

Now consider, in Figures 3 and 4, the part of the subspace $\mathcal{S}$ where $s > 1$. Depending on the actual values of $s$, $x_0$ and $p_0$, surface $E$ will enter this part of the subspace, for smaller values of $t$, through either surface $C$ or surface $D$ and exit the region, for larger $t$-values, through either surface $A$ or surface $B$. To determine through which of the surfaces $E$ actually enters and exists the part of the subspace under study, we establish the points at which the intersecting lines of the surfaces $A$ and $B$, and of the surfaces $C$ and $D$ meet. Writing $p_{VW}$ for the value of $p_0$ for which the lines $t_V$ and $t_W$ intersect, we find that

$$p_{AB} = \frac{s \cdot (x_0 - 1)}{x_0 - s}$$
$$p_{CD} = \frac{(1 - s) \cdot x_0}{x_0 - s}$$

From the above considerations, we now find that the intersection of $\mathcal{S}$ and $E$, which describes all combinations of values for $s$, $t$ and $r$ that capture a hyperbolic sensitivity function through $(x_0, p_0)$, is given by values for $s$ and $t$ that are related as described in Table 1 and $r = (x_0 - s) \cdot (p_0 - t)$. Note that for any point $(x_0, p_0)$ in the unit window and any value for $s$, the table provides an upper bound and a lower bound on the value of $t$ and hence, on the value of $r$. We illustrate the use of these bounds by means of an example.

**Example 3.1** We consider a parameter with an original value of $x_0 = 0.1$. In addition, we consider an output probability of interest with the original value $p_0 = 0.6$. Now suppose that $s = -2$. Then, from

- $p_{AB} = \dfrac{-2 \cdot (0.1 - 1)}{0.1 + 2} \approx 0.86 > p_0$, and

- $p_{CD} = \dfrac{(1 + 2) \cdot 0.1}{0.1 + 2} \approx 0.14 < p_0$

we find that the horizontal asymptote $t$ must lie within the interval $[t_B, t_D] \approx [-1.4, 1.93]$. We further find that $r = (x_0 - s) \cdot (p_0 - t)$ must lie within the interval $[-2.80, 4.20]$. These values for $t$ and $r$ now describe all possible sensitivity functions through $(0.1, 0.6)$ that have a vertical asymptote at $s = -2$. The bounding functions that enclose these possible sensitivity functions are depicted in Figure 5. □

Table 1: Upper and lower bounds on the range of possible $t$-values, for $s$ and $(x_0, p_0)$.

|  | $s < 0$ | $s > 1$ |
|---|---|---|
| $p_0 \geq p_{AB}$ | $t \geq t_A$ | $t \leq t_A$ |
| $p_0 \leq p_{AB}$ | $t \geq t_B$ | $t \leq t_B$ |
| $p_0 \leq p_{CD}$ | $t \leq t_C$ | $t \geq t_C$ |
| $p_0 \geq p_{CD}$ | $t \leq t_D$ | $t \geq t_D$ |

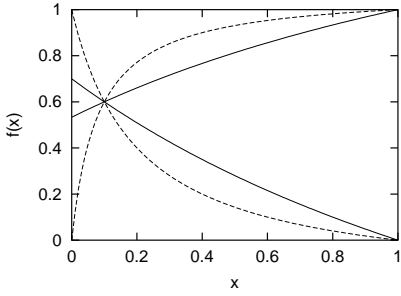

Figure 5: The bounds on all sensitivity functions through $(0.1, 0.6)$ with $s = -2$ (solid); the general bounds from Figure 2 are replicated for ease of reference (dashed).

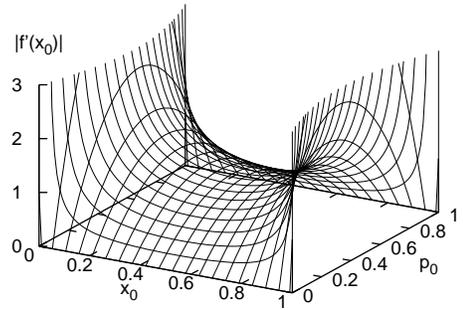

Figure 6: General bounds on the sensitivity value as a function of $x_0$ and $p_0$.

From the example we clearly see that additional knowledge of the value of the constant $s$ allows for further tightening the bounds on the range of possible sensitivity functions for a parameter under study.

### 3.2 THE FEASIBILITY OF ESTABLISHING $s$

The bounds on the values of the constants $t$ and $r$ established above, serve to describe all sensitivity functions through $(x_0, p_0)$ for a specific value of the constant $s$. We recall that the constant $s$ is defined as

$$s = -\frac{c_4}{c_3}, \text{ where } f_{\Pr(e)}(x) = c_3 \cdot x + c_4$$

The constant therefore is related to just the probability of the available evidence and is independent of the output probability of interest. Properties of the network under study may give an initial idea of the value of $s$. For example, if the distance in the network's digraph between the observed variables and the parameter that is being varied is quite large, then the probability of the evidence will be relatively insensitive to changes in the parameter's original value. The (absolute) value of $c_3$ will then be quite small and for rather likely evidence a relatively large (absolute) value of $s$ will result.

If required, the exact value of the constant $s$ can be computed in a very efficient way. From the method proposed by Kjaerulff & Van der Gaag for computing the constants of the sensitivity functions for a single output probability of interest with respect to all parameters in the network [6], we have that the constants $c_3$ and $c_4$ can be computed for *all* network parameters after just a single inward and a single outward propagation in the network's junction tree. One inward and one outward propagation therefore suffice to establish the value of $s$ for the given evidence, for *any* sensitivity function that we might be interested in, that is, for any parameter and any probability of interest. Note that actually computing all constants of all sensitivity functions requires an additional outward propagation for each output probability that we would like to consider and is therefore far more costly than just computing $s$.

To conclude, we would like to note that for establishing the value of $s$ we explicitly require the constants $c_3$ and $c_4$ and hence, the functional form of $f_{\Pr(e)}(x)$. Approaches that allow for computing $\Pr(e)$ in relation with $x$ without actually providing this functional form (see for example [5]) are therefore not suitable for our purposes.

## 4 APPLYING THE BOUNDS

In this section we use the evidence-dependent bounds established in the previous section to gain more insight in sensitivity values and for locating vertices.

### 4.1 BOUNDING SENSITIVITY VALUES

We recall that the sensitivity value of a parameter for an output probability of interest is the absolute value of $f'(x_0)$. Previously, it was shown that the sensitivity value for any sensitivity function through $(x_0, p_0)$ is bounded by the constant $(p_0 \cdot (1 - p_0))/(x_0 \cdot (1 - x_0))$ [2, 8]; Figure 6 depicts this general bound as a function of $x_0$ and $p_0$. Now that we have bounded the possible sensitivity functions through $(x_0, p_0)$ by using knowledge of the available evidence, we can also provide tighter evidence-dependent bounds on the sensitivity value of a parameter.

For any sensitivity function $f(x)$ through $(x_0, p_0)$ we have that

$$f'(x_0) = \frac{-r}{(x_0 - s)^2} = \frac{t - p_0}{x_0 - s}$$

By filling in the bounds $t_A, t_B, t_C$, and $t_D$ on the values of the constant $t$ as established in the previous section, we get the bounds on $f'(x_0)$ shown in Table 2. Taking the maximum of the absolute values of the upper bound and the lower bound, respectively, now gives us an upper bound on the sensitivity value that is dependent upon the value of $s$. An example of such an evidence-dependent upper bound is shown in the upper half of Figure 7. One of the more striking differences with Figure 6 is its asymmetry. To explain this difference, we observe that Figure 6 takes into account all possible values of $s$. For negative values of $s$, however, the hyperbola branch of the sensitivity function lies in the first or fourth quadrant. So, if a parameter is to have a large sensitivity value, its original value must be found among the smaller parameter values. For positive values of $s$, on the other hand, we will find the larger sensitivity values for the larger values of $x$. Knowledge of the value of $s$ thus serves to render the bound depicted in the upper part of Figure 7 asymmetric. From the example in the lower half of Figure 7, we note that for larger (absolute) values of $s$, the evidence-dependent bound starts to resemble the bound general found for linear sensitivity functions, which is reproduced in Figure 8 [8]. This tendency is readily explained from the observation that large values of $s$ can only be attained if the constant $c_3$ approaches zero (the constant $c_4$ can never be larger than one); for $c_3 = 0$, the sensitivity function is actually linear.

Table 2: Upper and lower bounds on $f'(x_0)$, for $s$ and $(x_0, p_0)$.

| | |
|---|---|
| $p_0 \geq p_{AB}$ | $f'(x_0) \geq \dfrac{-s \cdot (1 - p_0)}{-x_0 \cdot (x_0 - s)}$ |
| $p_0 \leq p_{AB}$ | $f'(x_0) \geq \dfrac{-p_0 \cdot (1 - s)}{(1 - x_0) \cdot (x_0 - s)}$ |
| $p_0 \leq p_{CD}$ | $f'(x_0) \leq \dfrac{p_0 \cdot s}{-x_0 \cdot (x_0 - s)}$ |
| $p_0 \geq p_{CD}$ | $f'(x_0) \leq \dfrac{(1 - p_0) \cdot (1 - s)}{(1 - x_0) \cdot (x_0 - s)}$ |

**Example 4.1** We consider again a parameter with an original value of $x_0 = 0.1$; we further consider an output probability of interest with the original value $p_0 = 0.6$. Now suppose that $s = -2$. We then find that

$$f'(x_0) \geq \frac{-0.6 \cdot (1 + 2)}{(1 - 0.1) \cdot (0.1 + 2)} \approx -0.95$$

and

$$f'(x_0) \leq \frac{(1 - 0.6) \cdot (1 + 2)}{(1 - 0.1) \cdot (0.1 + 2)} \approx 0.63$$

The sensitivity value of the parameter therefore can be no larger than 0.95. With the previously established general bound, we would have found an upper bound on the sensitivity value of $(p_0 \cdot (1 - p_0))/(x_0 \cdot (1 - x_0)) = 2.67$. Knowledge of the constant $s$ therefore serves to indicate that the parameter can have far less impact on the output probability than suggested by the general bound. □

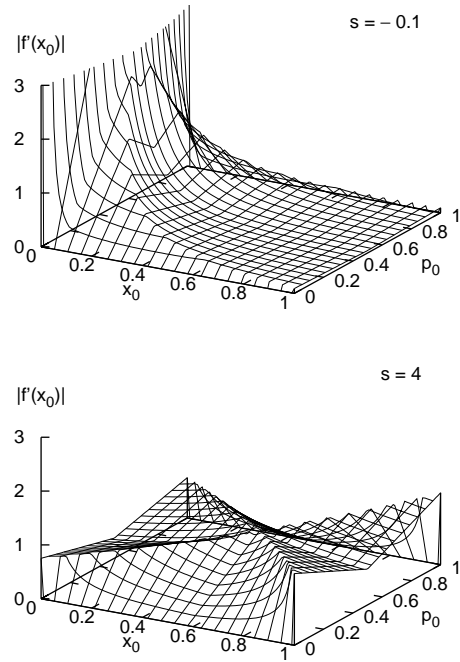

Figure 7: The bounds on the sensitivity value as a function of $x_0$ and $p_0$, for $s = -0.1$ and $s = 4$, respectively.

To conclude, we can derive the following simple general properties of sensitivity values: if $x_0 \geq p_0$ or $x_0 + p_0 \geq 1$, then any negative value for $s$ will result in a sensitivity value less than or equal to one; if $x_0 \leq p_0$ or $x_0 + p_0 \leq 1$, then any positive $s$-value will result in a sensitivity value less than or equal to one.

## 4.2 LOCATING VERTICES

We recall that the vertex of a hyperbola branch is the point where (the absolute value of) its first derivative equals one. If the vertex of a sensitivity function lies within the unit window, then values for the parameter under study with very high and very low sensitivity values may lie within a short distance of one another. For selecting parameters for further analysis, therefore, not only their sensitivity values, but also the vicinity of a vertex is of importance [9]. To gain insight in the location of the vertex of a sensitivity function, we recall that the vertices of the four possible hyperbola branches are found at the points $(x_v, y_v) = (s \pm \sqrt{|r|}, t \pm \sqrt{|r|})$. We now consider a hyperbolic sensitivity function that passes through a point $(x_0, p_0)$ in the unit window and address the question whether or not the vertex of the function is located in its vicinity. Note that since the vertex is a point itself on the sensitivity function, we need to study just its coordinate $x_v$.

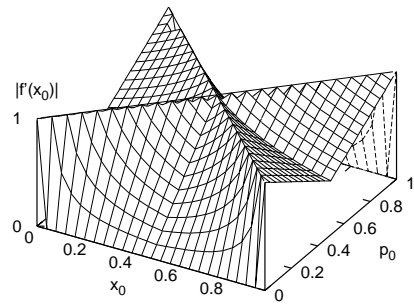

Figure 8: General bounds on the sensitivity value of linear sensitivity functions as a function of $x_0$ and $p_0$ [8].

Table 3: The ranges of $t$-values that, given $s$ and $(x_0, p_0)$, result in $x_v \in [\alpha, \beta]$.

| | |
|---|---|
| $p_{AB} \leq p_0 \leq p_{CD}$ | $t \in [t_{AC}^{\downarrow}, t_{AC}^{\uparrow}] \cap (t_1 \cup t_2)$ |
| $p_0 \geq \max\{p_{AB}, p_{CD}\}$ | $t \in [t_{AD}^{\downarrow}, t_{AD}^{\uparrow}] \cap (t_1 \cup t_2)$ |
| $p_0 \leq \min\{p_{AB}, p_{CD}\}$ | $t \in [t_{BC}^{\downarrow}, t_{BC}^{\uparrow}] \cap (t_1 \cup t_2)$ |
| $p_{CD} \leq p_0 \leq p_{AB}$ | $t \in [t_{BD}^{\downarrow}, t_{BD}^{\uparrow}] \cap (t_1 \cup t_2)$ |

We begin by addressing, for an arbitrary rectangular hyperbola branch that passes through the point $(x_0, p_0)$, whether or not the value $x_v$ of its vertex lies in a specific interval $[\alpha, \beta]$. By solving the constant $t$ from $x_v = s \pm \sqrt{|r|}$ with $r = (x_0 - s) \cdot (p_0 - t)$ for the four different quadrants, we find that $x_v \in [\alpha, \beta]$ iff

- $t \in \left[ p_0 + \dfrac{(s-\alpha)^2}{x_0 - s}, p_0 + \dfrac{(s-\beta)^2}{x_0 - s} \right]$, or

- $t \in \left[ p_0 - \dfrac{(s-\beta)^2}{x_0 - s}, p_0 - \dfrac{(s-\alpha)^2}{x_0 - s} \right]$

The first interval, denoted by $t_1$, is found for second- and fourth-quadrant branches; the second interval, denoted $t_2$, corresponds with first- and third-quadrant branches.

The two intervals established above pertain to all possible hyperbolas through the point $(x_0, p_0)$, but not yet to all possible hyperbolic sensitivity functions through $(x_0, p_0)$. To study the location of the vertices of hyperbolic sensitivity functions, therefore, the above intervals must be combined with the upper and lower bounds on the possible values for the constant $t$ from Table 1. Writing $t_{VW}^{\uparrow}$ and $t_{VW}^{\downarrow}$ for the maximum and the minimum of $t_V$ and $t_W$, $V, W \in \{A, B, C, D\}$, respectively, Table 3 gives the ranges of possible values for the constant $t$ that, for a specific $s$ and $(x_0, p_0)$, define a sensitivity function with $x_v \in [\alpha, \beta]$ for its vertex $(x_v, y_v)$.

**Example 4.2** We consider again a parameter with an original value of $x_0 = 0.1$ and an output probability of interest with the original value $p_0 = 0.6$. Suppose again that $s = -2$. We now are interested in whether or not the vertex $(x_v, y_v)$ of the real sensitivity function can be located within the unit window. We recall from Example 3.1 that $p_{CD} \leq p_0 = 0.6 \leq p_{AB}$, $t_B = -1.4$ and $t_D = 1.93$. In addition, we find with $\alpha = 0$ and $\beta = 1$ that

- $t_1 = [0.6 + \dfrac{4}{0.1+2}, 0.6 + \dfrac{(-2-1)^2}{0.1+2}] \approx [2.50, 4.89]$, and

- $t_2 = [0.6 - \dfrac{(-2-1)^2}{0.1+2}, 0.6 - \dfrac{4}{0.1+2}] \approx [-3.69, -1.30]$

The vertex of any hyperbolic sensitivity function through $(x_0, p_0)$ with a vertical asymptote at $-2$, therefore lies within the unit window iff

$$t \in [-1.40, 1.93] \cap ([2.50, 4.89] \cup [-3.69, -1.30])$$

With $s = -2$ and $t \in [-1.40, -1.30]$, we have that $r \in [4.0, 4.2]$; as a result, $x_v$ lies between $0$ and $0.05$, and $y_v$ between $0.65$ and $0.70$. If the true sensitivity function through $(x_0, p_0)$ with $s = -2$ has a vertex within the unit window, therefore, it must lie to the northwest of $(x_0, p_0)$; the horizontal distance to the vertex then is just between $0.05$ and $0.10$. We conclude that, although the sensitivity value of the parameter is at most $0.95$, variation to smaller values than $x_0$ might nonetheless induce a large change in the output probability of interest; variation to larger values, however, will certainly have no substantial effect. □

The analyses that were illustrated in the various examples in this section, can be performed for any point $(x_0, p_0)$ and any value of $s$. The theoretical results can subsequently be applied to the true points $(x_0, p_0)$ and true values of $s$ that are found in a real-life probabilistic network. Combinations that are potentially interesting due to the sensitivity value implied or the possible vicinity of a vertex can thus be distinguished from non-interesting combinations. Our analyses, for example, reveal the following behaviour, illustrated in Figure 9, as the vertical asymptote $x = s$ moves further away from the unit window:

- if $s$ is very close to zero (for example $s = -0.01$), then the vertex of the sensitivity function will also lie quite close to zero; as a result, very small values of the parameter $x$ under study will have large sensitivity values, while all other values for $x$ will have small sensitivity values. Similar observations apply if the constant $s$ is close to one.

- if $s$ moves slightly away from the window (for example to $s = 1.05$), then the sensitivity function starts to increase or decrease more gradually. A larger range of values for the parameter $x$ under study will then have large sensitivity values, although they will be smaller than the sensitivity values found for $s$ closer to zero or one.

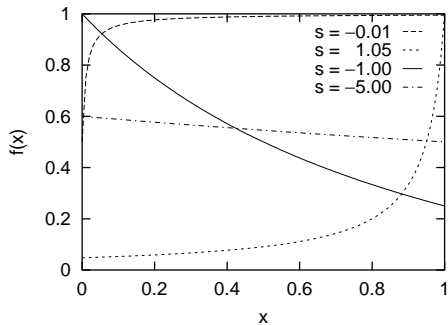

Figure 9: Examples of sensitivity functions with different values for $s$.

- if $s$ moves further away from the unit window (for example to $s = -1.00$ or $s = -5.00$), then the sensitivity function starts to flatten, resulting in small sensitivity values regardless of the value of the parameter under study.

## 5 CONCLUSIONS

Studying the effects of variation of any number of parameters on any number of output probabilities quickly becomes infeasible in practice. It is therefore useful to have generally applicable properties of sensitivity functions that can be exploited to distinguish between parameters for which further analysis is required and parameters that cannot show any substantial effect upon variation, without actually performing a full analysis. In this paper, we studied properties that build upon just the probability of the entered evidence. We demonstrated that these properties provide for establishing an evidence-dependent upper bound on the sensitivity value of a parameter that is tighter than the general bound established for networks in general. The properties further provide for bounding the region in which the vertex of a sensitivity function resides; the location of the vertex is important for identifying parameters with a low sensitivity value that may still have a large impact on the probability of interest for relatively small parameter shifts. We argued that establishing such evidence-dependent bounds requires just a single inward and a single outward propagation in a network's junction tree.

In many real-life applications of probabilistic networks, the outcome of interest is not a probability, but rather the most likely value of a variable of interest. For this type of outcome, the sensitivity value of a sensitivity function and the vicinity of a vertex no longer are appropriate for establishing the effect of parameter variation: a parameter with a large sensitivity value may upon variation not induce a change in the most likely outcome, while a parameter with a small sensitivity value may induce such a change for just a small deviation from its original assessment. To describe the sensitivities of this type of outcome, we introduced before the concept of *admissible deviation* [9]. This concept captures the extent to which a parameter can be varied without inducing a change in the most likely value for the variable of interest. Establishing properties of such admissible deviations that are independent of the network under study yet dependent of the available evidence, will be a challenging issue for further research.

## Acknowledgments

This research was (partly) supported by the Netherlands Organisation for Scientific Research (NWO).

## References

[1] E. Castillo, J.M. Gutiérrez, A.S. Hadi (1997). Sensitivity analysis in discrete Bayesian networks. *IEEE Transactions on Systems, Man, and Cybernetics*, vol. 27, pp. 412 – 423.

[2] H. Chan, A. Darwiche (2002). When do numbers really matter? *Journal of Artificial Intelligence Research*, vol. 17, pp. 265 – 287.

[3] H. Chan, A. Darwiche (2002). A distance measure for bounding probabilistic belief change. *Proceedings of the Eighteenth National Conference on Artificial Intelligence*, AAAI Press, Menlo Park, pp. 539 – 545.

[4] V.M.H. Coupé, L.C. van der Gaag (2002). Properties of sensitivity analysis of Bayesian belief networks. *Annals of Mathematics and Artificial Intelligence*, vol. 36, pp. 323 – 356.

[5] A. Darwiche (2000). A differential approach to inference in Bayesian networks. *Proceedings of the Sixteenth Conference on Uncertainty in Artificial Intelligence*, Morgan Kaufmann, San Francisco, pp. 123 – 132.

[6] U. Kjærulff, L.C. van der Gaag (2000). Making sensitivity analysis computationally efficient. *Proceedings of the Sixteenth Conference on Uncertainty in Artificial Intelligence*, Morgan Kaufmann, San Francisco, pp. 317 – 325.

[7] K.B. Laskey (1995). Sensitivity analysis for probability assessments in Bayesian networks. *IEEE Transactions on Systems, Man, and Cybernetics*, vol. 25, pp. 901 – 909.

[8] S. Renooij, L.C. van der Gaag (2004). Evidence-invariant sensitivity bounds. *Proceedings of the Twentieth Conference on Uncertainty in Artificial Intelligence*, AUAI Press, Arlington,VA, pp. 479-486.

[9] L.C. van der Gaag, S. Renooij (2001). Analysing sensitivity data from probabilistic networks. *Proceedings of the Seventeenth Conference on Uncertainty in Artificial Intelligence*, Morgan Kaufmann, San Francisco, pp. 530 – 537.